\documentclass[11pt,a4paper]{article}
\usepackage[hyperref]{emnlp2020}
\usepackage{times}
\usepackage{latexsym}

\usepackage{microtype}

\usepackage[utf8]{inputenc}
\usepackage{times}
\usepackage{url}
\usepackage{latexsym}
\usepackage{covington}
\usepackage{booktabs}
\usepackage{relsize}
\usepackage{todonotes}
\usepackage{xspace}
\usepackage{amsmath}
\usepackage{hyperref}
\usepackage{multirow}
\usepackage[all]{nowidow}
\usepackage{breakurl}

\newcommand{\ulmfit}{\textsc{ULMFiT}\xspace}    
\newcommand{\cnn}{\textsc{Cnn}\xspace}
\newcommand{\cnns}{\textsc{Cnns}\xspace}
\newcommand{\rnn}{\textsc{Rnn}\xspace}
\newcommand{\rnns}{\textsc{Rnns}\xspace}
\newcommand{\bow}{\textsc{Bow}\xspace}
\newcommand{\hcnn}{\textsc{Hcnn}\xspace}

\newcommand{\han}{\textsc{Han}\xspace}
\newcommand{\an}{\textsc{An}\xspace}
\newcommand{\mbert}{\textsc{mBert}\xspace}
\newcommand\F{{$\text{F}_1$}\xspace}

\aclfinalcopy{}

\title{A Systematic Comparison of Architectures \\ for Document-Level Sentiment Classification}

\author{Jeremy Barnes,
  Vinit Ravishankar,
  Lilja {\O}vrelid, and  
  Erik Velldal \\
  University of Oslo\\
  Department of Informatics\\
 {\tt \{jeremycb,vinitr,liljao,erikve\}@ifi.uio.no}}

\begin{document}
\maketitle
\begin{abstract}

Documents are composed of smaller pieces -- paragraphs, sentences, and tokens -- that have complex relationships between one another. Sentiment classification models that take into account the structure inherent in these documents have a theoretical advantage over those that do not. At the same time, transfer learning models have shown promise for document classification. However, these two paradigms have not been systematically compared and it is not clear under which circumstances one approach is better than the other. In this work we empirically compare flat, hierarchical, and transfer learning models for document-level sentiment classification in five languages. We show that the choice of model depends on data attributes and that hierarchical models perform similar to transfer learning, even in low-resource settings.
  
\end{abstract}

\section{Introduction}
\label{sec:intro}

The inherent structure found in documents -- paragraphs, sentences, and tokens -- and their inter-dependence is vital to document-level sentiment, as rhetorical devices and anaphora relationships disperse the sentiment signal across the various sub-components \cite{yang-cardie-2014-context}. This also means that not all sub-components contribute equally towards identifying the overall polarity of a document \cite{yu-hatzivassiloglou-2003-towards,pang-lee-2004-sentimental} and models that are able to take these relationships into account should theoretically perform better.

Recently, two divergent research directions have shown promise for document-classification: on the one hand, \emph{transfer learning} \cite{peters-etal-2018-deep,howard-ruder-2018-universal,devlin-etal-2019-bert} and on the other hand \emph{hierarchical modeling} \cite{Xia:Cho:16,Con:Sch:Bar:17,yang-etal-2016-hierarchical}. Transfer learning (in its current form) attempts to take advantage of large amounts of unlabeled text in order to improve contextualized representations of tokens, while ignoring the structure of documents. Hierarchical models, on the other hand, attempt to take document structure into account by first building up representations for sentences and then aggregating them to create document representations.

While the two approaches are complementary in the sense that one could use pretrained LMs for transfer-learning also for hierarchical models, we here focus on isolating their relative strengths and weaknesses. In this paper we empirically show that methods which explicitly incorporate the structure of documents outperform those that do not and further examine the influence of data characteristics such as document length and size of training data on the choice of architecture. 
Finally, we release the code to reproduce the results from our study. \footnote{\url{https://github.com/jerbarnes/hier_vs_transfer}}

\section{Background and related work}
\label{sec:background}

Document-level sentiment classification is a fundamental task in natural language processing and has a long tradition. Although there are document representation approaches which are more linguistically motivated, such as Rhetorical Structure Theory \cite{RST}, or centering \cite{grosz-sidner-1986-attention,grosz-etal-1995-centering}
these are not currently competitive with state-of-the-art approaches. In this section, we will review two current paradigms towards improving document-level classification: \emph{hierarchical models} and \emph{transfer learning}. 

\paragraph{Hierarchical models}

Hierarchical approaches to document classification aim to model the relationship between sub-components in a document by encoding first tokens, then sentences, and aggregating their representations in some way to create a full document representation which can be used for classification. The first work on hierarchical models for text classification was based on \cnns, either by stacking \cnn layers \cite{Zha:Zha:LeC:15,Con:Sch:Bar:17} or by using a single \rnn to aggregate the output of the convolutional layers \cite{Xia:Cho:16}. The performance of these models depends largely on the characteristics of the data, e.g., number of classes or dataset size, as these authors have conflicting findings on how many layers are optimal. Hierarchical models can also be based solely on \rnns. \citet{yang-etal-2016-hierarchical} propose a hierarchical model that uses an attention mechanism \cite{Bahdanau2015} over both sentence- and document-level GRU-based sequence encoders 
in order to attend to the most salient information, given the task. This model has shown promise for sentiment analysis and topic classification \cite{yang-etal-2016-hierarchical}, as well as classification of social media texts for e-health \cite{ive-etal-2018-hierarchical}. 


\paragraph{Transfer learning}

Transfer learning approaches, on the other hand, attempt to improve contextualized word representations, specifically by pretraining with a language modelling objective \cite{peters-etal-2018-deep, devlin-etal-2019-bert,Chang2019,wang-etal-2019-tell}. \citet{howard-ruder-2018-universal} pretrain a state-of-the-art LM \cite{merity2018regularizing} and introduce a number of improvements to the fine-tuning procedures. They demonstrate that this approach is able to make better use of later supervision. These approaches have shown promise for several document classification tasks, thanks largely to the availability of unannotated text and the size of the models used. However, to the best of our knowledge, these models have not been tested extensively on large documents.

\section{Data}

We perform experiments on document-level sentiment datasets in five languages: English, French, German, Japanese, and Norwegian. For the first four, we use the Amazon Customer Reviews datasets, a 5-class sentiment dataset with labels $L \in \{1, 2, 3, 4, 5\}$ stars.\footnote{\url{https://s3.amazonaws.com/amazon-reviews-pds/readme.html}} Although the full corpora are much larger, due to preprocessing requirements and in the interest of having similar sized data for all languages, we create a subcorpus $D$ by sampling 50,000 documents for each language without regarding domain, finally splitting these into test/dev/train splits of 35,000/5,000/10,000 documents. Each document is sentence split and tokenized using UDPipe \cite{udpipe:2017} and stored in CoNLL-U format. For Norwegian, we use the NoReC corpus 2.0, which is a 6-class task with labels $L \in \{1, 2, 3, 4, 5, 6\}$ ratings. It differs from version 1.0 \cite{Vel:Ovr:Ber:18} in that it has more training examples. Table~\ref{table:data} shows the statistics for each dataset.

\begin{table}[]
    \centering
    \resizebox{.47\textwidth}{!}{%
    \begin{tabular}{@{}lrcrrrr@{}}
    \toprule
    & $|D|$ & $|L|$ & T. & S. & T. / S. & $|V|$\\
    \midrule
    Fr &  50k & 5 & 81 & 6.3 & 12.9 & 100k\\ 
    De &  50k & 5 & 77  & 3.9 & 20.1 & 156k \\ 
    En & 50k & 5 & 114  & 8.0 & 14.3 & 109k\\
    Ja & 50k & 5 & 365 & 13.2 & 27.7 & 251k\\
    No & 43k & 6 & 463 & 27.8 & 16.7 & 564k \\
    \bottomrule
    \end{tabular}
    }
    \caption{Statistics ($|D|$ = number of documents, $|L|$ = number of labels, T. = average number of tokens per document, S. = average number of sentences per document, T./S. = average sentence length in tokens, $|V|$ = vocabulary size) for sentiment datasets.}
    \label{table:data}
\end{table}


\section{Experimental Setup}

The main research questions we seek to address in this section are: for document-level sentiment classification, are there systematic performance differences between flat, hierarchical, and LM-pretrained models, and do any of these approaches offer consistent improvements? Further, we investigate how performance is affected by several relevant data characteristics. Besides testing on the original data, we also test on documents where the order of the sentences is shuffled (\emph{shuffled}), to determine sensitivity to sentence order.

\subsection{Models}
We start by briefly summarizing the architectures. For all flat and hierarchical models, we use 200-dimensional randomly initialized embeddings that are updated during training, implemented in AllenNLP \cite{Gardner2017AllenNLP}. The flat and hierarchical models use randomly initialized word (sentence) embeddings that are updated during training. Unless stated otherwise, all models are trained for 50 epochs with a patience of 5 with Adam, and the best model is chosen using accuracy on the dev set.

\paragraph{\bow: } We train a linear SVM implemented in sklearn \cite{scikit-learn} on bag-of-words vectors learned from the train set and tune the $C$ parameter on the development set. 

\paragraph{\cnn: } \cnns are known to be strong baselines for document-level classification \cite{kim-2014-convolutional}. We implement a \cnn with filter sizes $F \in \{2,3,4,5\}$, with 50 filters per size and max pooling before a fully connected layer with a hidden size of 50. 

\paragraph{\an: } We use bidirectional Gated Recurrent Units \cite[GRU]{cho-etal-2014-learning} with dot product attention as a document encoder and a hidden size of 50.

\paragraph{Hierarchical \cnn (\hcnn):}
The Hierarchical \cnn uses filters $F_{sent} \in \{2,3,4,5\}$ and $F_{doc} \in \{2,3\}$, with 50 filters per size and max pooling before a fully connected layer to create sentence and document-level representations respectively. The output size for both layers is 50 dimensions. 

\paragraph{Hierarchical Attention Network (\han):}

Hierarchical Attention Networks \cite{yang-etal-2016-hierarchical} have shown promise for document-level tasks. We use Gated Recurrent Units \cite[GRUs]{cho-etal-2014-learning} as our encoders, with dot product attention. 

\paragraph{Universal Language Model Fine-Tuning (\ulmfit): } We use the AWD-LSTM architecture \cite{merity2018regularizing} and pretrain on Wikipedia data (or Common Crawl in the case of Norwegian) taken from the CONLL 2017 shared task \cite{conll-2017-conll}. The data was sentence and word tokenized using UDPipe \cite{udpipe:2017} and we perform no further preprocessing steps. 
We use between 14 and 18.7 million tokens (for No and Ja respectively) to pretrain the language model and choose the best model after pretraining for 100 epochs as determined by perplexity on the development set.


We then fine-tune the language models on the target domain, using slanted triangular learning rate schedule and finally fine-tune the models to the sentiment task using discriminative training proposed in \newcite{howard-ruder-2018-universal}. All experiments were performed using fastai \cite{howard2018fastai}.

\paragraph{Multilingual BERT (\mbert): } The multi-lingual BERT model is a transformer model \cite{Vaswani2017} which has been pretrained on masked-language and next sentence prediction tasks \cite{devlin-etal-2019-bert}. We use the cased model which was trained on Wikipedia dumps from 104 languages. We use the \textsc{[CLS]} token for prediction and finetune the model for 20 epochs, and test the model with the best performance on the dev set.

\begin{table}[]
    \centering
    
    \newcommand{\red}[1]{\textcolor{red}{#1}}
    \newcommand{\blue}[1]{\textcolor{blue}{#1}}
    \resizebox{.47\textwidth}{!}{%
    \begin{tabular}{@{}llrrrrrr@{}}
    \toprule
     & & En & No & Fr & De & Ja & Avg.\\
    \midrule
    \multirow{5}{*}{\rotatebox{90}{Flat}} &
    \bow & 71.1 & 55.6 & 63.3 & 73.2 & 60.6 & 64.2 \\
    &\cnn & 68.1 & 52.8 & 63.4 & 72.4 & 61.1 & 63.6 \\
    &{ }{ }{ }\emph{shuffled}  & \red{6.0} & \red{6.1} & \red{9.3} & \red{11.8} & \red{8.2} & \red{8.3}\\
    &\an & 71.9 & 58.1 & 63.9 & 74.3 & 62.1 & 66.1\\
    &{ }{ }{ }\emph{shuffled} & \red{0.1} & \red{1.8} & \blue{0.1} & 0.0 & \red{0.6} & \red{0.5}\\
    \midrule
    \multirow{4}{*}{\rotatebox{90}{Hier.}} &
    \hcnn & 68.3 & 55.7 & 61.9 & 71.2 & 61.0 & 63.6\\
    &{ }{ }{ }\emph{shuffled}  & \blue{0.3} & \red{0.1} & \red{0.2} & \blue{0.1} & \red{0.2} & 0.0 \\
    &\han & 72.1 & \textbf{61.4}$^{\dagger}$  & 63.2 & 73.2 & 61.9 & \textbf{66.4}\\
    &{ }{ }{ }\emph{shuffled}  & \red{0.6} & \red{1.9} & \red{0.4} & 0.0 & \red{0.3} & \red{0.7} \\
    \midrule
    \multirow{4}{*}{\rotatebox{90}{Transfer}} &
    \ulmfit & 69.4 & 55.7 & 60.8 & 69.3 & \textbf{63.7}$^{\dagger}$ & 63.8\\
    &{ }{ }{ }\emph{shuffled}  & \blue{0.1} & \red{15.8} & \red{0.3} & 0.0 & \red{0.4} & \red{3.3}\\
    &mBERT & \textbf{73.4}$^{\dagger}$  & 54.4 & \textbf{65.6}$^{\dagger}$ & \textbf{74.6} &  56.3 & 64.9 \\
    &{ }{ }{ }\emph{shuffled} & \red{0.7} & \red{8.0} & \red{0.5} & \red{0.8} & \blue{0.4} & \red{2.0}\\
    \bottomrule
    \end{tabular}
    }
    \caption{Accuracy of flat (\bow, \an, \cnn), hierarchical (\hcnn, \han), and transfer (\ulmfit, \mbert) models on document-level sentiment datasets ($\dagger$ denotes statistically significant difference (p $< 0.01$) when compared to the second best model. Bold numbers are the \textbf{best model} for each dataset, red numbers indicate \red{poorer performance} (in percentage points) when the sentence order is shuffled, while blue numbers indicate \blue{better} results when shuffled.}
    \label{table:results}
\end{table}

\subsection{Results}

Table \ref{table:results} shows the accuracy (\F results are similar) of the seven models for all five languages and their average, as well as statistical significance\footnote{We calculate statistical significance with approximate randomization testing \cite{yeh-2000-accurate} with 10000 runs.}). The \bow model performs well across all experiments, achieving an average 64.2 accuracy, and ties \han on the German dataset (73.2). The \cnn performs worse than the \bow across all experiments except Japanese (an average loss of 1.2 percentage points (pp)). \an has the second best overall performance (66.4).

Regarding the hierarchical models, the \hcnn performs on par with the flat \cnn (avg. 63.6), while the \han model is the best on Norwegian (61.4) and the best overall (avg. 66.4). This seems to indicate that while it is useful to explicitly model hierarchical structure using the \han model, the \hcnn is not as well suited to the task.

\ulmfit performs better than \bow on Norwegian and Japanese (0.1 / 3.1 pp), but 0.4 pp worse overall. 
\mbert is the best model on English, French, and German (73.4, 65.6, 74.6 respectively), but performs poorly on Norwegian (54.4) and Japanese (56.3). 

Shuffling the test data hurt 22 of 30 experiments and had the highest impact on the Norwegian and Japanese data -- an average drop of 5.6 pp -- both of which have the longest documents. The \cnn seems most sensitive to sentence order, losing 8.3 pp overall, followed by \ulmfit 3.3 and \mbert 2.0. Surprisingly, the hierarchical models are more robust to changes in sentence order.

\section{Analysis}
\label{analysis}

In this section we analyze what effects the number of training examples and document length has on the classifiers.

\subsection{Simulated low-resource settings}

In this section we compare transfer and hierarchical models in a simulated low-resource setting, where there are only a few training examples.  
\emph{A priori}, one might expect that transfer learning 
should perform better, given the use of additional unlabeled data.

\begin{figure}
    \centering
    \includegraphics[width=.5\textwidth]{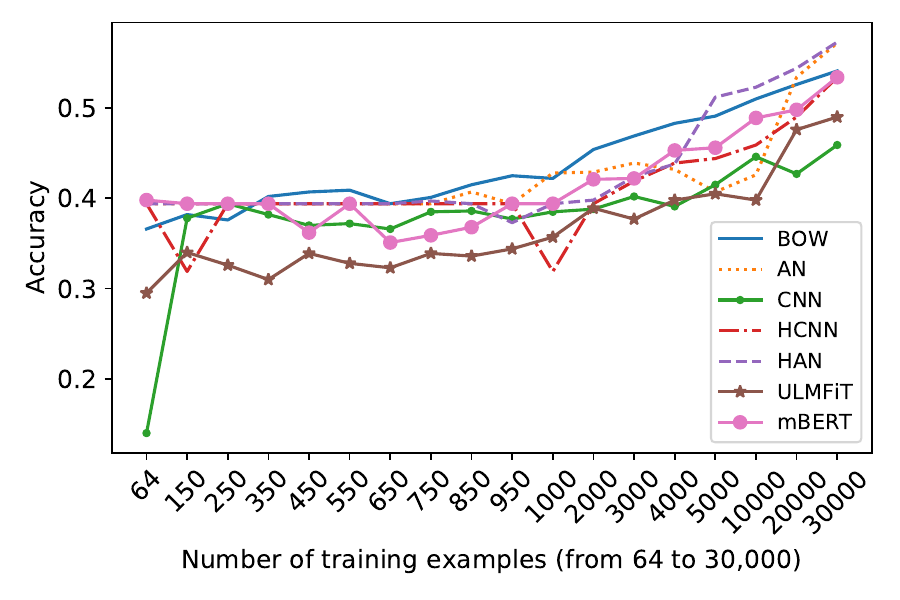}
    \caption{Accuracy on Norwegian test set with increasing number of training examples.}
    \label{fig:learningcurves}
\end{figure}

We compute learning curves by training models on NoReC (same results on the other data) using between 64 
to 30,000 labeled documents. Development and test data are kept the same. Figure \ref{fig:learningcurves} indicates that the models have the nearly same relative ranking with as few as 64 training examples as they do with 30,000, with the exception of \mbert. Surprisingly, even for low data scenarios, hierarchical modeling is as beneficial as transfer learning.

\subsection{Effect of document length}

Although most documents in the datasets are multi-sentence, they are not all of the same length. Therefore, we ask ourselves: \emph{do any of these models perform significantly better than other models on shorter/longer documents?}

\begin{figure}
    \centering
    \includegraphics[width=.5\textwidth]{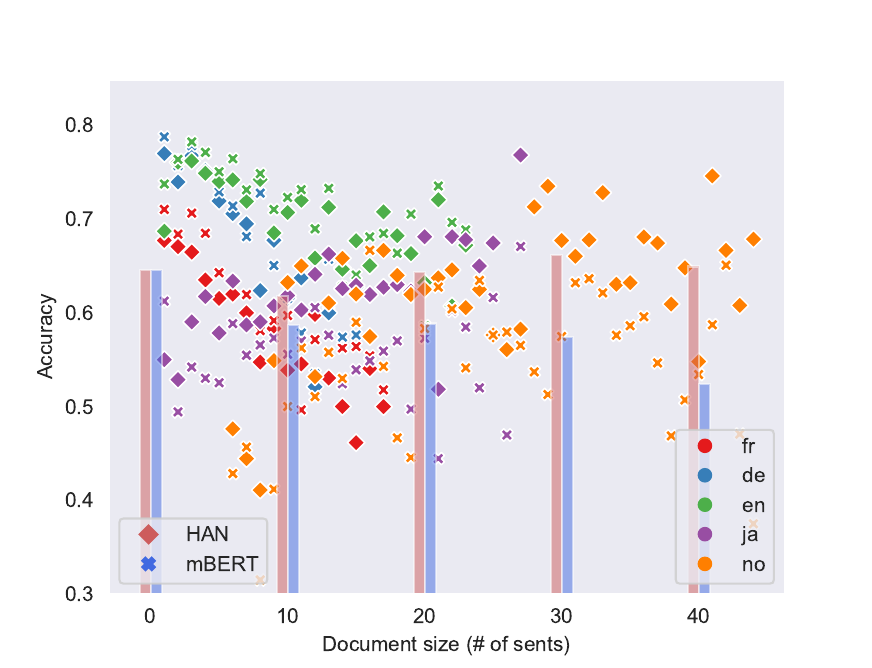}
    \caption{\han outperforms \mbert across different document lengths as determined by number of sentences per document. The scatter plot shows the accuracy on individual sentence counts (each color represents a dataset, and the markers represent the two models) while the bar plots show the mean accuracy of \han and \mbert respectively on 5 sentence-length bins.}
    \label{fig:sentlength}
\end{figure}

Figure \ref{fig:sentlength} shows the accuracy of \han and \mbert across the five languages on test documents, where the x-axis denotes document lengths (from 1 to 50), keeping those lengths that have more than 25 examples in order to avoid spurious results.
Although \mbert performs better on short documents, \han performs much better than \mbert on longer documents ($|d| > 10$), with Pearson ranked correlation of 0.41 (p $<$ 0.01). This trend also holds when comparing \han and \an. Hierarchical modeling, therefore, increases robustness to document length.

\section{Conclusion}
\label{sec:summary}

We have compared flat, hierarchical, and transfer learning models for document-level sentiment classification for five different languages and have shown that the best model depends on characteristics of your data. We also found that hierarchical models perform similar to transfer learning approaches even in low-resource scenarios, contrary to expectation.

%
\section*{Acknowledgements}
This work has been carried out as part of the SANT project (Sentiment Analysis for Norwegian Text), funded by the Research Council of Norway (grant number 270908).

\bibliographystyle{acl_natbib}
\interlinepenalty=10000
%
\bibliography{lit}

\end{document}